\documentclass[letterpaper, 10 pt, journal, twoside]{ieeetran} 
\usepackage{graphicx}
\usepackage{amsmath}
\usepackage{bbding}
\usepackage{pifont}
\usepackage{wasysym}
\usepackage{amssymb}
\usepackage{hyperref}
\usepackage{threeparttable} 
\usepackage{cite} 
\usepackage{booktabs}
\usepackage{xcolor}
\usepackage{multirow}
\usepackage{blindtext}
\usepackage{bm}
\usepackage{subcaption}
\usepackage[linesnumbered, ruled]{algorithm2e}

\SetKwInput{KwInput}{Input}
\SetKwInput{KwOutput}{Output}

\usepackage{cite}

\IEEEoverridecommandlockouts                              
\usepackage{geometry}
\usepackage{afterpage}
\usepackage{mwe}
\title{csBoundary: City-scale Road-boundary Detection in Aerial Images for High-definition Maps}
\author{Zhenhua Xu, \IEEEmembership{Student Member, IEEE}, 
Yuxuan Liu, \IEEEmembership{Student Member, IEEE},\\ Lu Gan, \IEEEmembership{Student Member, IEEE},  Xiangcheng Hu, \IEEEmembership{Student Member, IEEE}, Yuxiang Sun, \IEEEmembership{Member, IEEE}, \\Ming Liu, \IEEEmembership{Senior Member, IEEE}, and Lujia Wang, \IEEEmembership{Member, IEEE}%

\thanks{Manuscript received Sep 9, 2021; Revised December 13, 2021; Accepted February 1, 2022. This work was supported by Zhongshan Municipal Science and Technology Bureau Fund, under project ZSST21EG06, Foshan-HKUST Industry-University-Research Cooperation Project, under Project No. FSUST20-SHCIRI06C, and Department of Science and Technology of Guangdong Province Fund, under Project No. GDST20EG54, awarded to Prof. Ming Liu. 
\textit{(Corresponding author:  Lujia Wang.)}
}
\thanks{Zhenhua Xu is with the Department of Computer Science and Engineering, The Hong Kong University of Science and Technology, Clear Water Bay, Kowloon, Hong Kong (email: zxubg@connect.ust.hk).}
\thanks{Yuxiang Sun is with the Department of Mechanical Engineering, The Hong Kong Polytechnic University, Hung Hom, Kowloon, Hong Kong (email: yx.sun@polyu.edu.hk, sun.yuxiang@outlook.com).}
\thanks{Yuxuan Liu, Lu Gan, Xiangcheng Hu, Ming Liu and Lujia Wang are with the Department of Electronic and Computer Engineering, The Hong Kong University of Science and Technology, Clear Water Bay, Kowloon, Hong Kong ((email: \{\texttt{yliuhb},\texttt{lganaa},\texttt{xhubd}\}\texttt{@connect.ust.hk},\{\texttt{eelium},\texttt{eewang}\}\texttt{\\ @ust.hk}).).}
\thanks{Digital Object Identifier (DOI): see top of this page.} } 
\IEEEaftertitletext{\vspace{-2\baselineskip}}
  
\begin{document}

\maketitle
\begin{abstract}
High-Definition (HD) maps can provide precise geometric and semantic information of static traffic environments for autonomous driving. Road-boundary is one important information presented in HD maps since it distinguishes between road areas and off-road areas, which can guide vehicles to drive within road areas. But it is labor-intensive to annotate road boundaries for HD maps at the city scale. To enable automatic HD map annotation, current work uses semantic segmentation or iterative graph growing for road-boundary detection. However, the former could not ensure topological correctness since it works at the pixel level, while the latter suffers from inefficiency and drifting issues. To provide a solution to the aforementioned problems, in this letter, we propose a novel system termed csBoundary to automatically detect road boundaries at the city scale for HD map annotation. Our network takes as input an aerial image patch, and directly infers the continuous road-boundary graph (i.e., vertices and edges) from this image. To generate the city-scale road-boundary graph, we stitch the obtained graphs from all the image patches. Our csBoundary is evaluated and compared on a public benchmark dataset. The results demonstrate our superiority. The project page is available at \url{https://sites.google.com/view/csboundary/}.          
              
\vspace{0.25cm}

\end{abstract}

\section{Introduction}
 
\IEEEPARstart{R}{oad} boundary is important for autonomous vehicles. It can distinguish road areas from off-road areas, so that vehicles could be constrained within safe regions and potential accidents could be avoided. Early work usually detects road boundaries with on-vehicle sensors, such as LiDAR and camera \cite{lu2020real, sun20193d}. However, robustly detecting road boundaries is challenging, since boundaries are long-and-thin and usually with irregular shapes. Moreover, occlusions often happen on real roads, which severely degrades the detection performance. To provide a solution to the aforementioned problems, high-definition maps (HD maps) have been widely used in existing autonomous driving systems. Recent progress in this area has witnessed several methods using aerial images to automatically annotate line-shaped objects (e.g., road-lane and road-boundary) in HD maps.


Typically, HD maps are hand-labeled from bird-eye-view (BEV) images, such as high-resolution aerial images, or overhead images from pre-built point-cloud maps. With the rapid development of aerial photography and remote sensing, high-resolution aerial images could be easily accessed all over the world. In addition, unlike pre-built point-cloud maps that are expensive to create and update, high-resolution aerial images are more cost-effective. In our previous work, we released a benchmark dataset of aerial images,  topo-boundary \cite{xu2021topo}, for road-boundary detection. With this dataset, we propose to automatically annotate a city-scale HD map of road boundaries in New York City (NYC) in this work.

As a kind of geographic information system (GIS), HD map has two primary ways to record spatial data: vector representation and raster representation. For line-shaped objects such as road curbs, vector representation (i.e., graph with vertices and edges) is usually adopted. Therefore, to automatically annotate the HD map of road boundaries, we need to obtain the graph of road boundaries. In real-world applications, since the aerial images usually cover a very large area (e.g., a whole city), we cannot directly produce the whole graph of the area due to the limitation of computation resources. Instead, we apply a sliding window to crop image patches and stitch the obtained graph of each patch into the final city-scale HD map. In this way, automatic HD maps annotation is divided into two sub-tasks: (1) predict the graph of road boundaries within an image patch, and (2) stitch the graph of different patches into a large city-scale graph as the final draft HD map.

Few past works have exactly the same scope as this work (i.e., automatically annotate the city-scale HD map of road boundaries from BEV aerial images), while they focus on related tasks, such as road-lane detection \cite{homayounfar2018hierarchical,homayounfar2019dagmapper,can2021structured}, road-network detection \cite{bastani2018roadtracer,tan2020vecroad,he2020sat2graph,etten2020city}, road-curb detection \cite{zhxu2021icurb,xu2021cp} and road-boundary detection \cite{liang2019convolutional,xu2021topo}. These works could be classified into three primary categories: segmentation-based methods, iterative-graph-growing methods, and graph-generation methods. Most early works on line-shaped object detection belong to segmentation-based methods \cite{mattyus2017deeproadmapper,batra2019improved}. They first predict the segmentation map of the target object and conduct post-processing algorithms to extract the final graph, such as skeletonization. Due to the poor topology correctness of segmentation-based methods, some recent work \cite{bastani2018roadtracer,tan2020vecroad,zhxu2021icurb,xu2021cp} iteratively grow the graph vertex-by-vertex in a sequential manner. Even though this category of methods presents much better topology correctness, they suffer from the drifting issue (i.e., error accumulation) and awful parallelization capability. To address the shortcomings of the aforementioned works, He \textit{et al.}\cite{he2020sat2graph} first proposed to directly generate the graph of line-shaped objects by using a carefully designed graph encoding scheme.

In this paper, we first predict the key vertices of the input aerial image, then predict the adjacency matrix of the obtained key vertices for graph edges. Since the length of the obtained key vertices is variant, in the past, RNN and iterative operations are utilized to handle various length input \cite{mi2021hdmapgen}. But RNN and iterative operations are not efficient and cannot make full use of long-term memories, thus a better approach for adjacency matrix prediction is required. Compared with RNN that requires sequential operations, transformer \cite{vaswani2017attention} can directly handle various length input and is easier to be parallelized. Considering the aforementioned characteristics of transformer, in this paper, we propose to use it for adjacency matrix prediction, so that the graph could be generated without neither complicated post-processing nor iterative steps. To the best of our knowledge, this is the first paper that makes use of transformer to predict graphs for automatic HD map annotation. The contributions of this work are summarized as follows:
\begin{enumerate}
    \item We propose a new approach to define keypoints of line-shaped objects for road-boundary graph vertex detection.
    \item We propose a novel adjacency matrix prediction network named attention for adjacency network (AfANet).
    \item We design a system named csBoundary for city-scale road-boundary HD map automatic annotation in aerial images.
\end{enumerate}

\section{Related Works}
\subsection{Segmentation-based methods}
Many early works on line-shaped object detection extract graphs by two-step segmentation-based methods \cite{mattyus2017deeproadmapper,batra2019improved,mnih2010learning,etten2020city}. They first predict the segmentation probabilistic map. Since segmentation maps are in the raster format, a series of post-processing is then conducted to refine the segmentation results and extract the graph by geometric techniques, such as binarization and skeletonization. Batra \textit{et al.} \cite{batra2019improved} made use of the orientation map to enhance the segmentation result of road networks, and trained another network to refine the segmentation result, which greatly improves the correctness of the final output. However, even with carefully-designed post-processing, this category of method still suffers from serious topology errors, such as incorrect disconnections and ghost connections.

\subsection{Iterative-graph-growing methods}
RoadTracer \cite{bastani2018roadtracer} is believed to be the first work that predicts the graph of line-shaped objects by iterative-graph-growing methods. The authors first manually selected several initial vertices of the road network. Then, starting from these initial vertices, a decision-making network was trained to predict the coordinate of the next vertex. In this way, the road-network graph was generated vertex-by-vertex through iterative graph growing. This method is also adapted to other tasks, such as road-boundary detection \cite{liang2019convolutional,xu2021topo} and road-curb detection \cite{zhxu2021icurb}. \cite{liang2019convolutional} could present satisfactory results on road-boundary graph prediction, but it only works on the highway with simple and clean scenarios. Our previous work \cite{xu2021topo} could achieve good detection performance, but it takes a huge amount of time for training and inference due to the inefficient iterative steps. Moreover, since the prediction error is accumulated with the growing graph, this category of method is difficult to be extended to city-scale tasks.

\subsection{graph-generation methods}
Directly predicting graphs from images is a challenging task, since graphs may have different numbers of vertices and the relationship between vertices (e.g., edges) is difficult to formulate. There are some past works utilizing vector fields to achieve graph prediction \cite{xue2019learning,he2020sat2graph,girard2021polygonal}. Xue \textit{et al.} \cite{xue2019learning} aimed to predict line segments of an image. They proposed a vector field named \textit{attraction field} which could be predicted by segmentation networks. Then the authors designed a decoding scheme to recover line segments from the predicted \textit{attraction field}. Similarly, \cite{girard2021polygonal} proposed a new vector field to predict the polygon of buildings in satellite images. \cite{he2020sat2graph} is believed to be the first work that detects line-shaped objects in BEV images of this category of methods. In this paper, each pixel of the input image was encoded by a 19-dimensional vector. Then the 19-dimensional encoding tensor was predicted by neural networks. Finally, the graph was decoded from the predicted encoding tensor by the proposed decoding algorithm. However, this method cannot distinguish edges with small included angles. Moreover, this category of methods heavily relies on the heuristic decoding algorithms, which limits their generalization ability.



\subsection{Transformer}
Transformer \cite{vaswani2017attention} has been widely applied in deep learning tasks in recent years. The main module of transformer is the self-attention layer, which could handle various length input. Compared to RNN that has been widely used in the past, transformer is much more efficient due to the good parallelization ability \cite{vaswani2017attention}. Transformer has been applied in graph neural networks (GNNs) \cite{yun2019graph}, but extracting graphs from images is not fully explored yet. 
To the best of our knowledge, this is the first work that uses transformer for automatic HD map annotation.

\section{The Proposed Method}
\begin{figure*}[t]
  \centering
    \includegraphics[width=\linewidth]{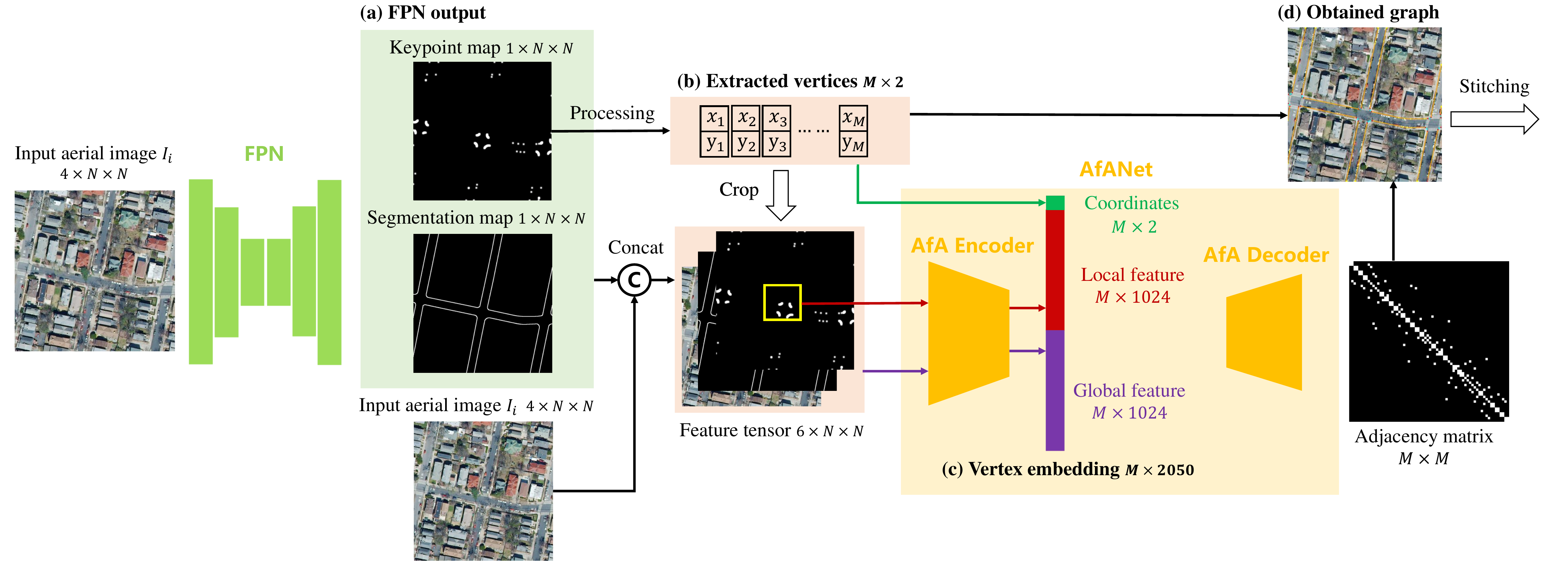}
  \caption{The system overview of csBoundary. (a) Taken as input a 4-channel aerial image $I_i$, we first predict the keypoint map and segmentation map of road boundaries by using FPN. Then these maps are concatenated into a new 6-channel feature tensor; (b) Based on the predicted keypoint map, a set of processing is conducted to extract the vertex coordinates of the graph of which the length is $M$; (c) For each extracted vertex, we crop a $L\times L$-sized ROI centering at the keypoint. Then the ROI is sent to an encoder network to calculate the local feature vector. Similarly, a global feature vector can be obtained. Then we concatenate feature vectors with the keypoint coordinate into the final embedding of the current vertex; (d) AfANet predicts the adjacency matrix of extracted vertices based on the vertex embeddings. Then, the road-boundary graph $G_i$ of $I_i$ is obtained based on vertices and the adjacency matrix. Finally, we stitch the graph of all aerial images $\{G_i\}_{i=1}^N$ into the final city-scale road-boundary graph $G$. For better visualization, only RGB channels are visualized for aerial images. Please zoom in for details.}
  \label{diagram}
\end{figure*}
\subsection{The method overview}
In this paper, we aim to solve the problem of automatically annotating city-scale road-boundary HD maps using aerial images. Suppose the input is a set of aerial image patches $\{I_i\}_{i=1}^N$ which covers a large area (e.g., a whole city), then the output should be a city-scale road-boundary graph $G=(V,E)$. Since the city-scale aerial images cover a very large area, the road-boundary graph could not be obtained directly due to the limitation of computation resources. Therefore, the problem is divided into two sub-problems: (1) how to detect the road-boundary graph $G_i$ in a single aerial image patch $I_i$ cropped by a sliding window; and (2) how to stitch the predicted graph of all patches $\{G_i\}_{i=1}^N$ into the final city-scale road-boundary graph $G$. The graph $G$ can be used as the draft HD map of road boundaries for autonomous driving. The pseudocode of our system is shown in Alg. \ref{alg1}, and the corresponding section ID is listed in the comment of each key step. Please refer to our supplementary document \cite{csBoundary} for more details.

\begin{algorithm}[!h]

\KwInput{A set of aerial image patches $A=\{I_i\}_{i=1}^N$}
\KwOutput{A city-scale undirected \& unweighted graph $G$}
\Begin{
$K_p \gets \emptyset$, $G_p \gets \emptyset$ \\
Image patch expansion \# \ref{expansion} \\
\While{$A$ not empty}{
        $I\gets A.pop()$\\
        $K \gets FPN(I)$ \# \ref{FPN} \\
        $K_p.push(K)$
   }
Keypoint segmentation map stitching \# \ref{stitching}\\
\While{$K_p$ not empty}{
        $K\gets K_p.pop()$ \\
        Extract graph vertices as $V$ \# \ref{vertex extraction} \\
        Predict edges by AfANet as $E$ \# \ref{AfANet}\\
        $G_p.push(G_i=(V,E))$\\
        }
Graph stitching \# \ref{stitching}\\
return $G$}

\caption{The proposed csBoundary}
\label{alg1}
\end{algorithm}

Like \cite{he2020sat2graph}, our csBoundary belongs to the graph-generation method and it predicts the road-boundary graph without heuristic post-processing or iterative operations. csBoundary first predicts two probabilistic maps by the feature pyramid network (FPN) \cite{lin2017feature}, including a keypoint map and a road-boundary segmentation map. These two maps are then concatenated with the input aerial image $I_i$ into a 6-D feature tensor. Based on the predicted keypoint map, we conduct a series of processing to find the local maximum, and extract the coordinates of graph vertices, whose length is denoted by $M$. To predict the adjacency matrix of vertices, inspired by the self-attention mechanism in transformer networks, we propose the attention for adjacency network (AfANet). Taken as input the 6-D feature tensor and coordinates of extracted graph vertices, AfANet directly outputs the adjacency matrix. Centering at each extracted vertex, we crop a $L\times L$-sized region of interest (ROI) on the 6-D feature tensor and calculate a 1024-length local feature vector by the AfANet encoder. Similarly, the whole 6-D feature tensor is sent to the encoder to obtain a 1024-length global feature vector. These two feature vectors are concatenated together with the coordinate of the current vertex as the final vertex embedding, whose length is 2050. After processing $M$ extracted vertices, we have $M$ 2050-length vertex embedding vectors. Then, AfANet predicts the adjacency matrix of graph vertices by the decoder network. Based on the predicted graph vertices and adjacency matrix, we can compute the graph $G_i$ of the input aerial image patch $I_i$. Finally, we stitch the graph of all patches $\{G_i\}^{N}_i$ into the final city-scale road-boundary graph. 

\begin{figure}[t]
  \centering
    \includegraphics[width=\linewidth]{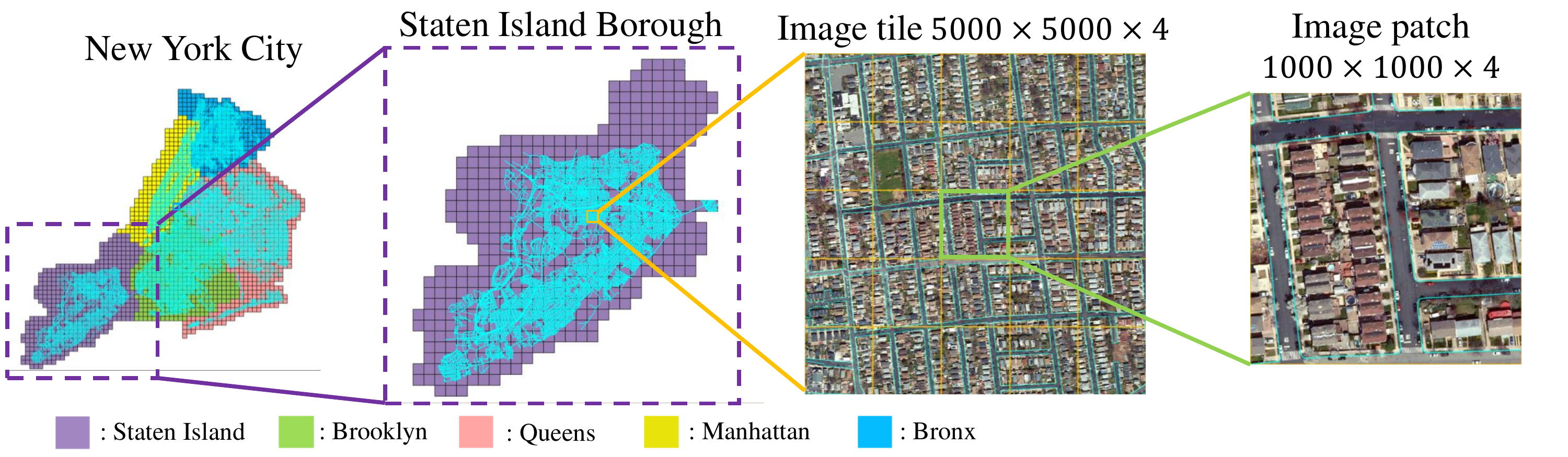}
  \caption{Visualization of data split in the original dataset. The aerial images of the dataset cover the whole NYC. There are 5 boroughs in NYC and they are illustrated in different colors. In each borough, there are a set of image tiles whose size is $5000\times 5000$. Considering the limited GPU memory, each image tile is further split into 25 $1000\times 1000$-sized image patches. There are no intersection areas between adjacent patches. }
  \label{fig:data crop}
\end{figure}
\subsection{Aerial image data split and expansion} \label{expansion}
\begin{figure}[t]
  \centering
    \includegraphics[width=\linewidth]{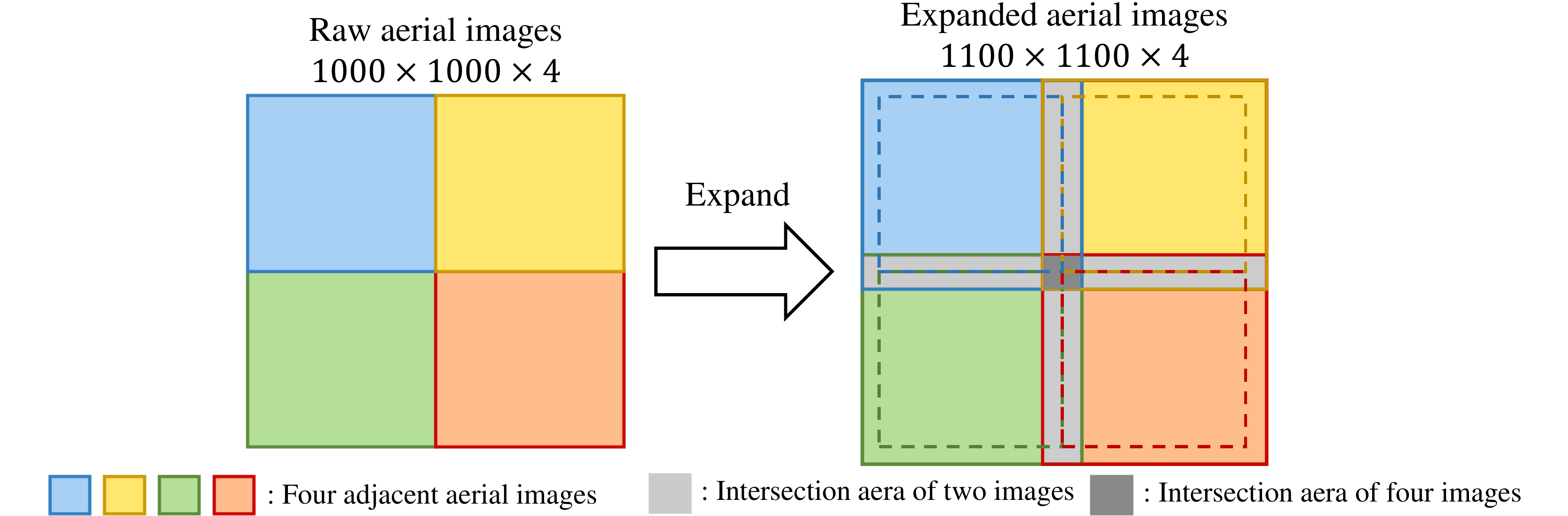}
  \caption{Visualization of image patch expansion. In this paper, all aerial image patches are expanded into $1100\times 1100$-size, which creates intersection areas between adjacent patches (light gray and dark gray areas). Dashed rectangles on the right represent the original image edges. These intersection areas will benefit the graph stitching process.}
  \label{fig:image expansion}
\end{figure}

In this paper, the aerial images are from the benchmark dataset released in our previous work \cite{xu2021topo}. In the dataset, there are 2,049 4-channel $5000\times 5000$-sized high-resolution aerial image tiles that cover 5 boroughs of the whole NYC. Due to the memory limitation of GPU devices, we split each tile into 25 $1000\times 1000$-sized image patches. The data split method is visualized in Fig. \ref{fig:data crop}. In our previous work, we did not consider graph stitching and removed some image patches based on proposed filtering rules. While in this paper, we keep all the image patches and follow the idea of \cite{etten2020city} for city-scale graph stitching. For each image patch, we expand its size to create overlapping areas between adjacent image patches. The overlapping areas are critical to the stitching process. More details will be discussed in the following subsections. The visualization for the image expansion is shown in Fig. \ref{fig:image expansion}.

\subsection{FPN and keypoint map} \label{FPN}
Feature pyramid network (FPN) \cite{lin2017feature} is widely used in the past works on line-shaped object detection \cite{liang2019convolutional,homayounfar2019dagmapper,zhxu2021icurb} due to its great ability to capture multi-scale image features. In csBoundary, FPN is utilized for keypoint map and segmentation map prediction. The keypoint map can detect keypoints of road boundaries, and some keypoints will be treated as vertices of the output graph. The segmentation map is used to detect foreground road-boundary pixels. 

\begin{figure}[t]
  \centering
    \includegraphics[width=\linewidth]{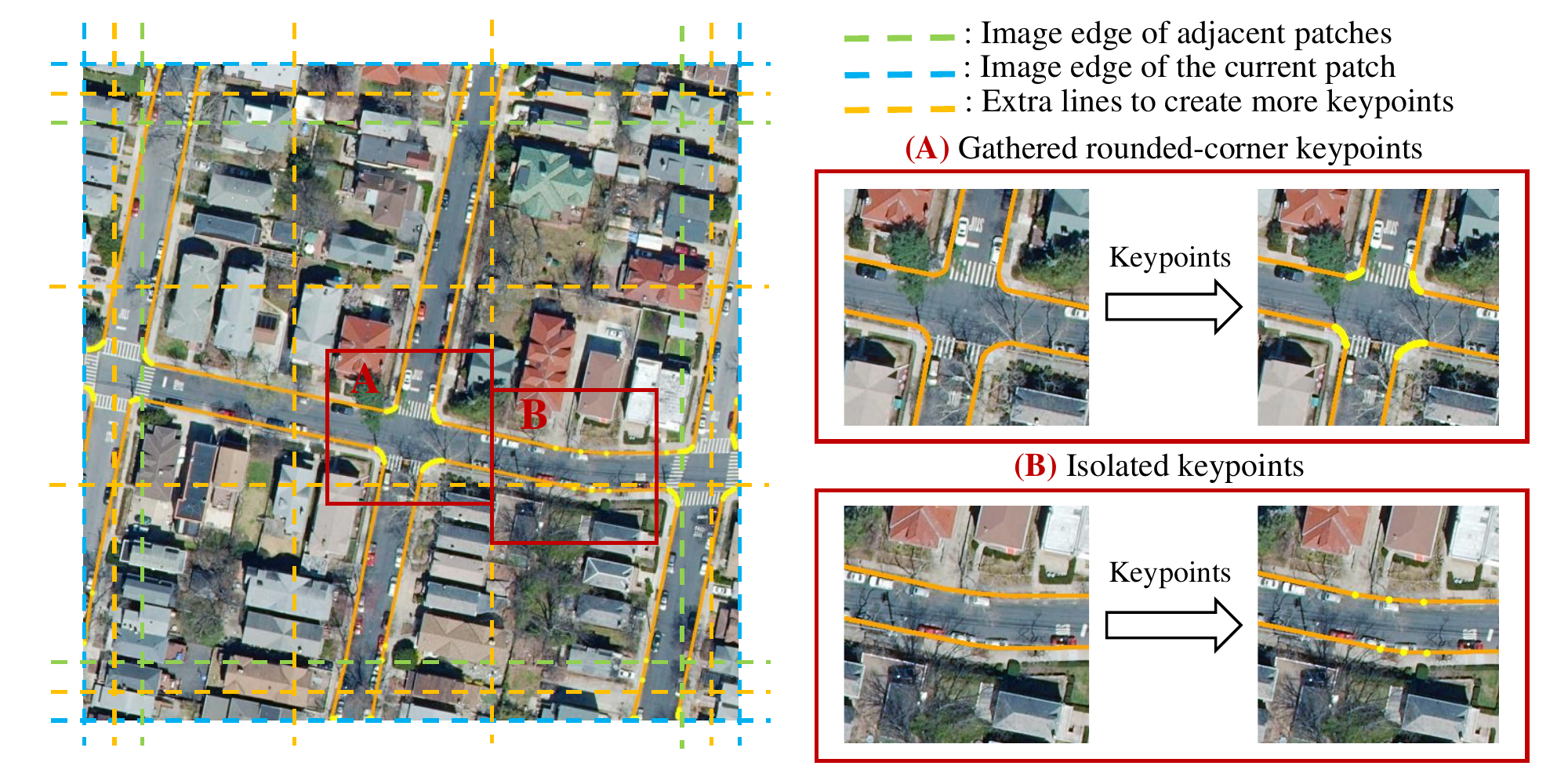}
  \caption{Demonstration of keypoints. There are generally two types of keypoints. (1) Intersection keypoints. This category of keypoints is the intersection points of road boundaries and manually defined lines, such as the edge of the current patch (blue dash line), the edge of adjacent patches (green dash line) and extra lines (orange dash line); (2) Corner keypoints. They could be gathered keypoints at rounded-corners (yellow points in region A) or isolated keypoints locating at curve road boundaries with sharp corners or gradual curvature changes (yellow points in region B). All keypoints are uniquely defined.  }
  \label{fig: keypoint}
\end{figure}

Unlike human pose estimation \cite{sun2019deep} and road network detection \cite{he2020sat2graph} that usually have clearly defined unique keypoints (e.g., joints for human skeleton and crossroad for road network), road boundaries are usually polylines without branches. Therefore, it is hard to find unique keypoints with clear semantic meanings. To provide a solution to conquer this, we make use of the orientation map \cite{batra2019improved} that records the direction vector of each pixel, and define pixels whose orientation has large enough differences with adjacent pixels as keypoints. In short, pixels where the road-boundary curvature is large enough are treated as keypoints. In addition, within each image patch, we define the intersection points of the road-boundary and image edges as keypoints for graph stitching. Sometimes the keypoints defined by the aforementioned methods may be too sparse, thus we add extra lines to create more intersection points as auxiliary keypoints. Examples of keypoints are visualized in Fig. \ref{fig: keypoint}.

\subsection{Vertex extraction} \label{vertex extraction}
\begin{figure}[t]
  \centering
    \includegraphics[width=\linewidth]{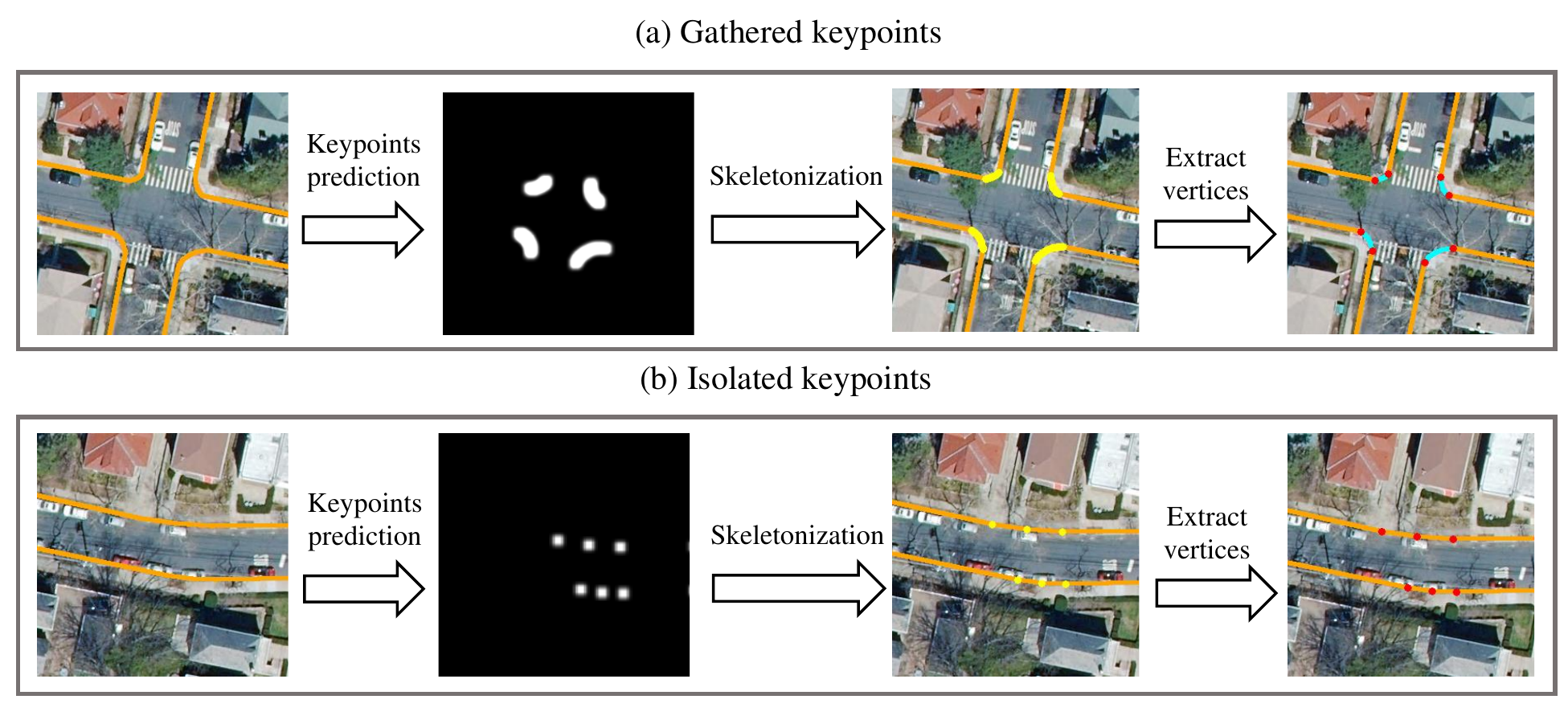}
  \caption{Vertex extraction pipeline. Yellow points denote the skeleton of the predicted keypoint map, and red points represent the extracted graph vertices. (a) For gathered keypoints (i.e., whose skeleton is a curved line segment), we only keep two endpoints of the skeleton and add them into the graph vertex set (red points), while other points are not added (cyan line). In the final graph, we will connect the endpoints by the cyan line; (b) For isolated keypoints, we directly calculate its skeleton and add the center point of the skeleton into the graph vertex set. }
  \label{fig: extraction}
\end{figure}

Graph vertices are extracted from the predicted keypoint map by finding local peaks. After obtaining the predicted keypoint map, we first find its skeleton. Then, for isolated keypoints whose corresponding skeleton instances are short, we directly use the center of the skeleton as graph vertices. While for rounded-corner keypoints that many points gather together, the skeleton will be curved line segments, then we only add endpoints of the curved line segments into the graph vertex set. In the final graph, the rounded-corner vertices are connected by corresponding skeletons directly without adjacency matrix prediction. The vertex extraction pipeline is shown in Fig. \ref{fig: extraction}.

\subsection{Adjacency matrix prediction} \label{AfANet}
After graph vertex extraction, the connection relationship between vertices (i.e., edges) should be predicted. In past works, this is usually done by heuristic algorithms that decode edges from carefully designed vector field maps \cite{he2020sat2graph,girard2021polygonal}. To further enhance the effectiveness and efficiency of graph edge prediction, inspired by transformer and self-attention mechanism, we propose the attention for adjacency net (AfANet) to predict the adjacency matrix of the graph.

First, for each extracted graph vertex, we calculate a vertex embedding by the AfANet encoder. Then we put the embedding vector of all vertices into the AfANet decoder and obtain the adjacency matrix.
\subsubsection{AfANet encoder}
The vertex embedding is produced by a multi-layer convolutional encoder network. Each vertex embedding is of 2050-length, which is concatenated by a 1024-length global feature, a 1024-length local feature and the normalized 2-D coordinates of the corresponding vertex. The global feature is obtained by directly passing the 6-D feature tensor through the encoder network. For the local feature, we crop a $L\times L$-sized ROI ($L$ is 64 in our experiment) on the 6-D feature tensor and send it to another branch of the encoder network. Suppose $M$ vertices are extracted, then we will have a $M\times 2050$-sized feature tensor containing the information of all vertices by the encoder network.

\subsubsection{AfANet decoder}
 \begin{figure}[t]
        \begin{subfigure}[t]{0.17\textwidth}
            \includegraphics[width=\textwidth]{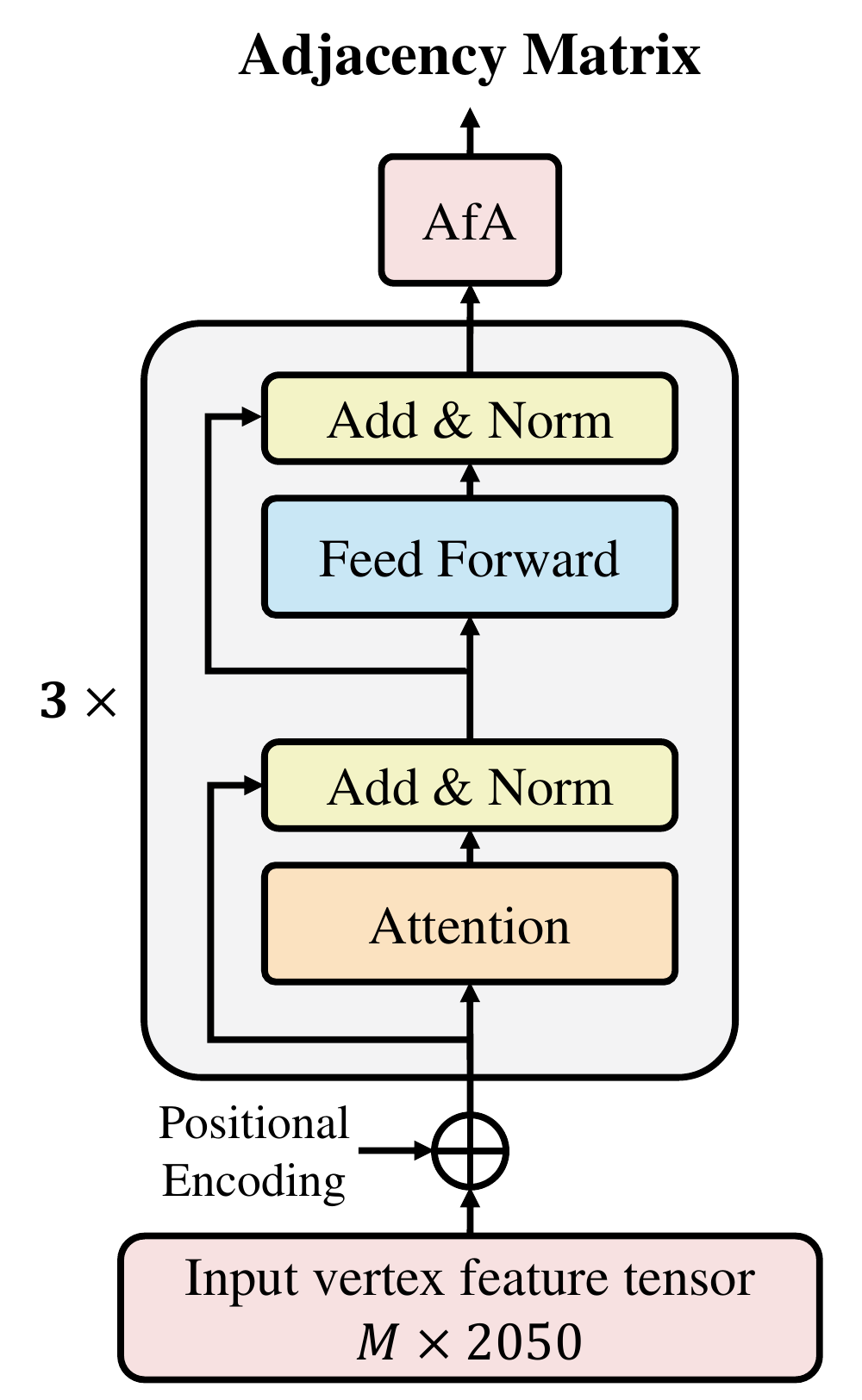}
            \caption{AfANet decoder}
        \end{subfigure}
        \begin{subfigure}[t]{0.17\textwidth}
          \includegraphics[width=\textwidth]{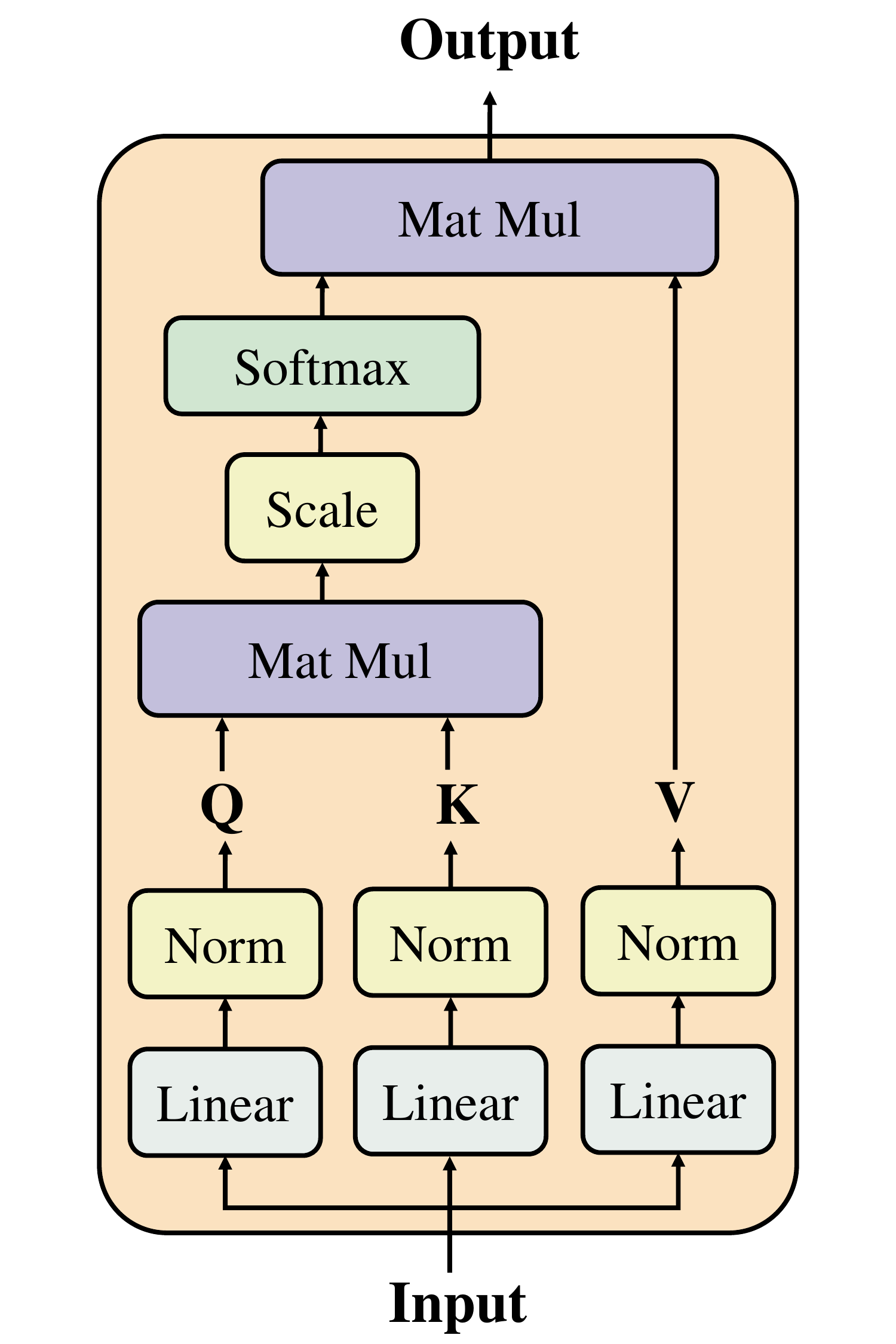}
            \caption{Attention module}
        \end{subfigure}
        \begin{subfigure}[t]{0.12\textwidth}
          \includegraphics[width=\textwidth]{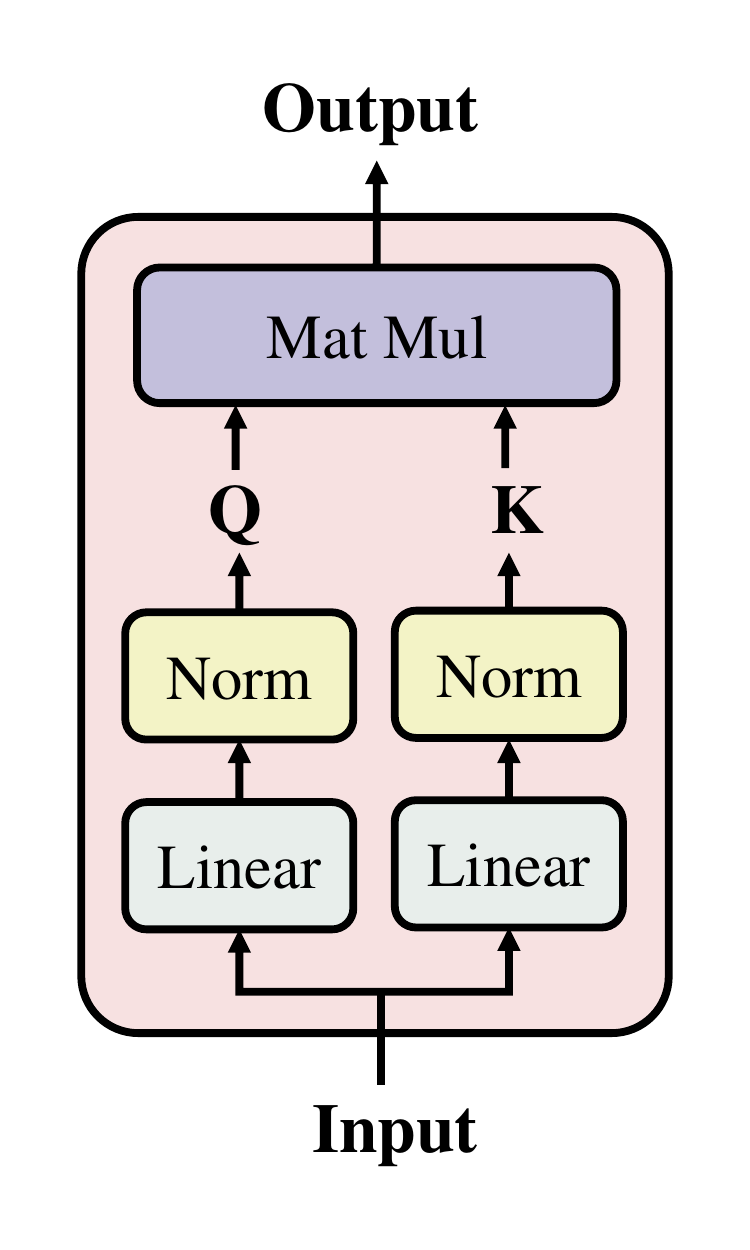}
            \caption{AfA module}
        \end{subfigure}
    
    \caption{The visualization of AfANet decoder. Taken $M\times 2050$-sized vertex embedding tensor as input, AfANet decoder predicts the $M\times M$-sized adjacency matrix of extracted graph vertices. (a) The general network structure of the AfANet decoder; (b) The modified attention module; (c) The AfA module.}
    \label{fig: AfANet}
\end{figure}
AfANet decoder is inspired by the self-attention mechanism. It can handle various length input and predict the adjacency matrix of the input directly. Suppose the size of the input is $M\times d$, then the shape of output is $M\times M$. The attention module of AfANet decoder is modified from the original self-attention module \cite{vaswani2017attention}:
\begin{equation}
    \text{Attention}(\textbf{Q},\textbf{K},\textbf{V}) = \text{softmax}(norm(\textbf{Q})\cdot norm(\textbf{K}))\cdot \textbf{V}
\end{equation}
The AfA module only utilizes Q(Query) and K(Key), and outputs the dot-product attention map as the adjacency matrix:
\begin{equation}
    \text{AfA}(\textbf{Q},\textbf{K}) = norm(\textbf{Q})\cdot norm(\textbf{K})
\end{equation}
The structure of the AfANet decoder is visualized in Fig. \ref{fig: AfANet}. 

\subsection{Graph stitching} \label{stitching}
\begin{figure}[t]
  \centering
    \includegraphics[width=\linewidth]{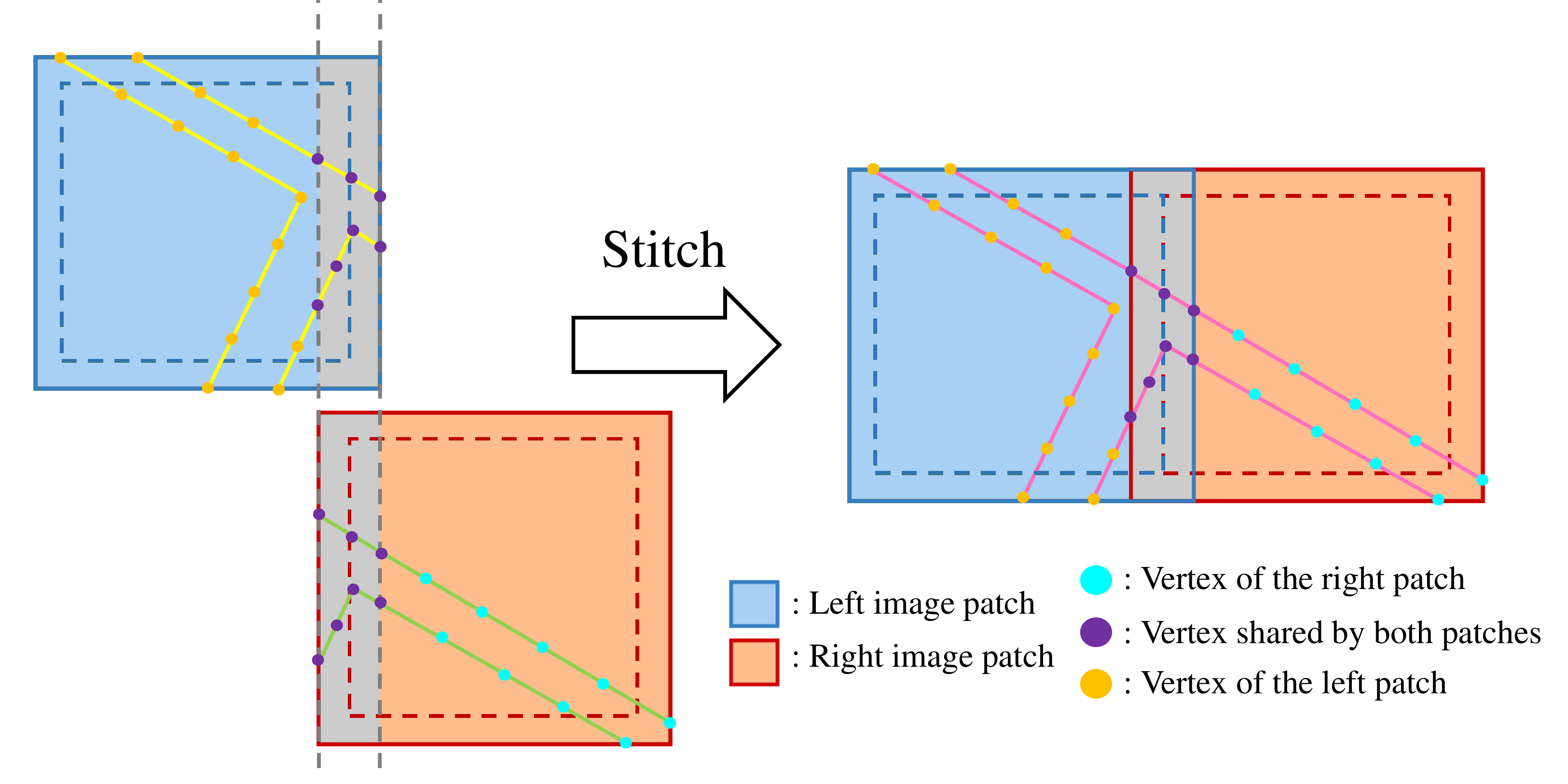}
  \caption{Demonstration of graph stitching. The blue image patch is adjacent to the red one (blue patch on the left and red patch on the right). For better visualization, they are placed vertically. The gray areas are the intersection areas. The graph of the blue patch (yellow edge and orange vertex) and the red patch (green edge and cyan vertex) are predicted separately. Purple vertices are shared by both patches. These two patches could be stitched together easily by connecting exclusive vertices of two patches with the shared vertices.}
  \label{fig: graph stitch}
\end{figure}

Following the Broad Area Satellite Imagery Semantic Segmentation (BASISS) method \cite{etten2020city}, before extracting vertices from keypoint maps, we stitch predicted keypoint maps by averaging the intersection areas (e.g., for an intersection area of two keypoint maps, its value is the average of corresponding areas of these two keypoint maps). In this way, the intersection area of adjacent predicted keypoint maps will be exactly the same. Then, we extract vertices from the keypoint map of both image patches, and there will be some shared vertices within the intersection area. The graph of the two adjacent image patches could be stitched together easily by connecting exclusive vertices of two patches with the shared vertices. An example demonstrating the graph stitching process is shown in Fig. \ref{fig: graph stitch}.

\section{Experimental Results and Discussions}
\subsection{Dataset}
We use the dataset released from our previous work topo-boundary \cite{xu2021topo}. But different from topo-boundary, in this paper, we aim to solve the city-scale road-boundary graph detection problem, while topo-boundary only focuses on patch-scale detection. Topo-boundary also removes some image patches by proposed filtering rules, such as hard patches with complicated scenarios. In this paper, we consider all patches without filtering.

There are 2,049 4-channel $5000\times 5000$-sized aerial image tiles in the dataset. We split each tile into 25 $1000\times 1000$-sized image patches considering the limited memory resources of GPU devices. Then for city-scale graph stitching, we expand each image patch into $1100\times 1100$-size. We split the dataset by borough (Manhattan, Brooklyn, Queens, Bronx, and Staten Island), which is visualized in Fig. \ref{fig:data crop}. In our experiments, the Staten Island borough is for testing, while other boroughs are for training and validation.


\subsection{Implementation details}
In our experiments, we first train the FPN for 10 epochs with the learning rate as 0.001 and decay rate as $10^{-4}$. Then we extract graph vertices based on predicted keypoint maps. To enhance the performance of AfANet, we first pre-train the network with ground-truth graph vertices, and then train AfANet with the graph vertices extracted from predicted keypoint maps for 20 epochs. The adjacency matrix label can be calculated by using graph vertices and ground-truth road-boundary binary label maps. During inference, we run an extra graph stitching step. We conduct experiments on a PC with an i7-8700K CPU and an RTX3090 GPU. 
For better evaluation and comparison, in our experiments, we provide the patch-scale evaluation results (i.e., graph within a single patch) which are the same as topo-boundary, and the city-scale evaluation results (i.e., graph after the stitching step).

\subsection{Evaluation metrics}
In our experiments, we have 5 metrics for evaluation, including 3 pixel-level metrics (i.e., Precision, Recall and F1-score) following our previous work \cite{xu2021topo}, and two topology-level metrics, i.e., Average path length similarity (APLS) \cite{van2018spacenet} and too long/too short (TLTS) similarity \cite{wegner2013higher}. These metrics are sufficient to provide a comprehensive and fair comparison for different methods.

\subsubsection{Pixel-level metrics}
Precision, Recall and F1-score are three relaxed metrics to measure the correctness of the predicted graph at pixel level. We first rasterize the predicted road-boundary graph (i.e., convert vector graph to raster image), and denote obtained foreground pixels as $P=\{p_i\}_{i=1}^{N_p}$. Similarly, we rasterize the ground-truth road-boundary graph as $Q=\{q_i\}_{j=1}^{N_q}$. Suppose the relax ratio is $\tau$. Then we have
\begin{equation}
    \begin{aligned}
        &Precision = \frac{|\{
        p | d(p,Q)<\tau,\forall p\in P
        \}|}{|P|},\\
        &Recall = \frac{|\{
        q | d(q,P)<\tau,\forall q\in Q
        \}|}{|Q|},\\
        &F1\text{-}score =\frac{2Precision\cdot Recall}{Precision+Recall},
\end{aligned}
\end{equation}
where $|\cdot|$ represents the number of elements of a set, and $d(e,S)$ calculates the shortest Euclidean distance between an element $e$ and a set $S$. Relax ratio $\tau$ could reflect the level of error tolerance. In our experiments, we show the results by setting $\tau$ as 2, 5 and 10 pixels, respectively.


\subsubsection{APLS and TLTS}
In many past works, APLS and TLTS are utilized to measure the topology correctness of the obtained graph. Let $G$ denote the ground-truth graph and $P$ denote the predicted graph. Then we randomly select two vertices $\{g_1,g_2\}$ from $G$, and calculate the shortest path between $g_1$ and $g_2$ as $l(g_1,g_2)$. Finally, we find the two corresponding vertices $\{p_1,p_2\}$ in $P$ and calculate $l(p_1,p_2)$. Then the APLS score of this vertex pair is 
\begin{equation}
    APLS = 1 - min(1,\frac{|l(g_1,g_2)-l(p_1,p_2)|}{l(g_1,g_2)})
\end{equation}

After sampling a certain number of vertex pairs, the final APLS score is the mean of APLS of all vertex pairs. TLTS also relies on randomly sampled vertex pairs and shortest path calculation. Define an error tolerance threshold $\phi$ (default value is 0.05 in our experiment), if the difference between $l(g_1,g_2)$ and $l(p_1,p_2)$ is larger enough, i.e.,
\begin{equation}
    |l(g_1,g_2)-l(p_1,p_2)| > l(g_1,g_2) \cdot \phi,
\end{equation}
then this vertex pair is said to be \textit{too long or too short}. TLTS is the ratio of vertex pairs that are not \textit{too long or too short}.


\subsection{Comparative results}
 \begin{figure*}[t]
    \newcommand{\picvisi}[1]{925137}
    \newcommand{\picvisii}[1]{935137}
 \centering
    \begin{subfigure}[t]{0.195\textwidth}
        \begin{subfigure}[t]{\textwidth}
            \includegraphics[width=\textwidth]{img/gt_crop_\picvisi{_}.pdf}
        \end{subfigure}\vspace{.6ex}
        \begin{subfigure}[t]{\textwidth}
            \includegraphics[width=\textwidth]{img/gt_crop_\picvisii{_}.pdf}
        \end{subfigure}\vspace{.6ex}
        \caption{Ground-truth}
        \label{fig_qualitative_1st}
    \end{subfigure}
    \begin{subfigure}[t]{0.195\textwidth}
        \begin{subfigure}[t]{\textwidth}
            \includegraphics[width=\textwidth]{img/or_crop_\picvisi{_}.pdf}
        \end{subfigure}\vspace{.6ex}
        \begin{subfigure}[t]{\textwidth}
            \includegraphics[width=\textwidth]{img/or_crop_\picvisii{_}.pdf}
        \end{subfigure}\vspace{.6ex}
        
        \caption{OrientationRefine \cite{batra2019improved}}
        \label{fig_qualitative_1st}
    \end{subfigure}
    \begin{subfigure}[t]{0.195\textwidth}
        \begin{subfigure}[t]{\textwidth}
            \includegraphics[width=\textwidth]{img/ei_crop_\picvisi{_}.pdf}
        \end{subfigure}\vspace{.6ex}
        \begin{subfigure}[t]{\textwidth}
            \includegraphics[width=\textwidth]{img/ei_crop_\picvisii{_}.pdf}
        \end{subfigure}\vspace{.6ex}
        
        \caption{Enhanced-iCurb \cite{xu2021topo}}
        \label{fig_qualitative_1st}
    \end{subfigure}
    \begin{subfigure}[t]{0.195\textwidth}
        \begin{subfigure}[t]{\textwidth}
            \includegraphics[width=\textwidth]{img/sg_crop_\picvisi{_}.pdf}
        \end{subfigure}\vspace{.6ex}
        \begin{subfigure}[t]{\textwidth}
            \includegraphics[width=\textwidth]{img/sg_crop_\picvisii{_}.pdf}
        \end{subfigure}\vspace{.6ex}
        
        \caption{Sat2Graph \cite{he2020sat2graph}}
        \label{fig_qualitative_1st}
    \end{subfigure}
    \begin{subfigure}[t]{0.195\textwidth}
        \begin{subfigure}[t]{\textwidth}
            \includegraphics[width=\textwidth]{img/cs_crop_\picvisi{_}.pdf}
        \end{subfigure}\vspace{.6ex}
        \begin{subfigure}[t]{\textwidth}
            \includegraphics[width=\textwidth]{img/cs_crop_\picvisii{_}.pdf}
        \end{subfigure}\vspace{.6ex}
        
        \caption{Ours}
        \label{fig_qualitative_1st}
    \end{subfigure}
    \caption{Qualitative visualization. Each subfigure is of $2000\times2000$-sized. (a) The ground truth (cyan lines); (b) Results of  OrientationRefine (green lines); 
    (c) Graph obtained by enhanced-iCurb (orange lines); 
    (d) Graph obtained by Sat2Graph (orange lines). It cannot present reasonable results since its encoding scheme cannot be well adapted to our task; (e) Graph obtained by csBoundary. Yellow points are normal vertices, red points are rounded-corner vertices, orange lines are normal edges and cyan lines are edges connecting corresponding rounded-corner vertices. Please zoom in for details. }
    \label{fig_qualitative}
\end{figure*}

In this section, we evaluate csBoundary together with the other three baseline models that belong to different categories of methods. The city-scale evaluation results are shown in Tab. \ref{tab_comparative} and patch-scale evaluation results are listed in Tab. \ref{tab_comparative_patch}. The average time usage is shown in Tab. \ref{tab_time}.  
\begin{itemize}
    \item OrientationRefine (ICCV2019)\cite{batra2019improved}: This baseline is a typical segmentation-based work. It first predicts the segmentation map of road networks and then corrects the segmentation results by another refinement network.
    
    \item Enhanced-iCurb (RA-L2021)\cite{xu2021topo}: This baseline is the state-of-the-art iterative-graph-growing work. Starting from initial vertices, it iteratively generates the road-boundary graph vertex by vertex.
    
    \item Sat2Graph (ECCV2020)\cite{he2020sat2graph}: This baseline is believed to be the first work that can directly predict the graph of line-shaped objects from the graph generation perspective. After extracting keypoints, Sat2Graph can obtain graph edges by decoding the predicted vector field map.
\end{itemize}
\begin{table*}[th] 
\setlength{\abovecaptionskip}{0pt} 
\setlength{\belowcaptionskip}{0pt} 
\renewcommand\arraystretch{1.0} 
\renewcommand\tabcolsep{10.4pt} 
\centering 
\begin{threeparttable}
\caption{The patch-scale quantitative comparative results. The best results are highlighted in bold font. 
For all the metrics, larger values indicate better performance.
}
\begin{tabular}{@{}c c c c c c c c c c c c c c c c@{}}
\toprule
\multirow{3}{*}{Methods}& \multicolumn{3}{c}{Precision $\uparrow$} & \multicolumn{3}{c}{Recall $\uparrow$} & \multicolumn{3}{c}{F1-score $\uparrow$}& \multirow{3}{*}{APLS $\uparrow$}& \multirow{3}{*}{TLTS $\uparrow$} \\ 
\cmidrule(l){2-4} \cmidrule(l){5-7} \cmidrule(l){8-10} 
&   2.0 &  5.0 &  10.0 &   2.0 &  5.0 &  10.0
&    2.0 &  5.0 &  10.0 & \\
\midrule
OrientationRefine \cite{batra2019improved} &\textbf{0.507}&\textbf{0.805}&0.865&0.408&0.650&0.706&\textbf{0.439}&0.699&0.753& 0.462& 0.437\\ 
Enhanced-iCurb \cite{xu2021topo} & 0.458&0.744&0.825&\textbf{0.446}&\textbf{0.713}&0.785&0.433&\textbf{0.704}&0.778&0.707& 0.669
 \\
 Sat2Graph \cite{he2020sat2graph} & 0.343& 0.624& 0.757& 0.139& 0.268& 0.326& 0.163& 0.322& 0.397& 0.150& 0.134
 \\
csBoundary 
& 0.333& 0.704&\textbf{0.877}& 0.300& 0.650&\textbf{0.808}& 0.306& 0.682& \textbf{0.825}& \textbf{0.734}& \textbf{0.690}
\\
\bottomrule 
\label{tab_comparative_patch}
\end{tabular} 
\end{threeparttable}
\end{table*}

\begin{table*}[th] 
\setlength{\abovecaptionskip}{0pt} 
\setlength{\belowcaptionskip}{0pt} 
\renewcommand\arraystretch{1.0} 
\renewcommand\tabcolsep{10.4pt} 
\centering 
\begin{threeparttable}
\caption{The city-scale quantitative comparative results. The best results are highlighted in bold font. For all the metrics, larger values indicate better performance.
} 
\begin{tabular}{@{}c c c c c c c c c c c c c c c c@{}}
\toprule
\multirow{3}{*}{Methods}& \multicolumn{3}{c}{Precision $\uparrow$} & \multicolumn{3}{c}{Recall $\uparrow$} & \multicolumn{3}{c}{F1-score $\uparrow$}& \multirow{3}{*}{APLS $\uparrow$}& \multirow{3}{*}{TLTS $\uparrow$} \\ 
\cmidrule(l){2-4} \cmidrule(l){5-7} \cmidrule(l){8-10} 
&   2.0 &  5.0 &  10.0 &   2.0 &  5.0 &  10.0
&    2.0 &  5.0 &  10.0 & \\
\midrule
OrientationRefine \cite{batra2019improved} &0.517&\textbf{0.816}&\textbf{0.868}&0.352&0.551&0.589&0.408&0.637&0.678& 0.235& 0.219\\ 
Enhanced-iCurb \cite{xu2021topo} & 0.412&0.695&0.785&\textbf{0.412}&\textbf{0.671}&\textbf{0.749}&\textbf{0.410}&\textbf{0.678}&0.760&0.299& 0.279
 \\
  Sat2Graph \cite{he2020sat2graph} &\textbf{0.460}&0.484&0.604&0.128&0.240&0.293&0.159&0.304&0.374&0.037& 0.030
 \\
csBoundary &
0.309&0.659&0.830&0.291&0.600&0.738&0.297&0.652&\textbf{0.772}&\textbf{0.376}& \textbf{0.343}
\\
\bottomrule 
\label{tab_comparative}
\end{tabular} 
\end{threeparttable}
\end{table*}
\begin{table}[t] 
\setlength{\abovecaptionskip}{0pt} 
\setlength{\belowcaptionskip}{0pt} 
\renewcommand\arraystretch{1.0} 
\renewcommand\tabcolsep{4.3pt} 
\centering 
\begin{threeparttable}
\caption{The time consumption of the methods. We report the average time taken for one epoch.} 
\begin{tabular}{c c c c c}
\toprule
 & OrientationRefine & Enhanced-iCurb & Sat2Graph & csBoundary \\
\midrule
Training&4.98h&86.01h&4.33h&2.47h\\
Inference&0.76h&9.10h&1.45h&0.82h\\
\bottomrule 
\label{tab_time}
\end{tabular} 
\end{threeparttable}
\end{table}
\begin{table*}[h] 
\setlength{\abovecaptionskip}{0pt} 
\setlength{\belowcaptionskip}{0pt} 
\renewcommand\arraystretch{1.0} 
\renewcommand\tabcolsep{10.4pt} 
\centering 
\begin{threeparttable}
\caption{The quantitative results of city-scale ablation studies. The best results are highlighted in bold font. For all the metrics, larger values indicate better performance.} 
\begin{tabular}{@{}c c c c c c c c c c c c c c c c@{}}
\toprule
\multirow{3}{*}{Methods}& \multicolumn{3}{c}{Precision $\uparrow$} & \multicolumn{3}{c}{Recall $\uparrow$} & \multicolumn{3}{c}{F1-score $\uparrow$}& \multirow{3}{*}{APLS $\uparrow$}& \multirow{3}{*}{TLTS $\uparrow$} \\ 
\cmidrule(l){2-4} \cmidrule(l){5-7} \cmidrule(l){8-10} 
&   2.0 &  5.0 &  10.0 &   2.0 &  5.0 &  10.0
&    2.0 &  5.0 &  10.0 & \\
\midrule
Without global feature & \textbf{0.375}&\textbf{0.699}&0.818&0.254&0.474&0.545&0.286&0.538&0.623&0.311& 0.307\\ 
Without local feature & 0.270& 0.628 & 0.813&0.221&0.505&0.657&0.240&0.569&0.716&0.345&0.319
 \\
csBoundary & 0.309&0.659&\textbf{0.830}&\textbf{0.291}&\textbf{0.600}&\textbf{0.738}&\textbf{0.297}&\textbf{0.652}&\textbf{0.772}&\textbf{0.376}& \textbf{0.343}
 \\
\bottomrule 
\label{tab_ablation}
\end{tabular} 
\end{threeparttable}
\end{table*}
From Tab. \ref{tab_comparative_patch} and Tab. \ref{tab_comparative}, it is found that pixel-level metric scores are similar for patch-scale evaluation and city-scale evaluation, since pixel-level metrics focus on locality, which is not greatly affected by the graph stitching process. However, APLS and TLTS of the city-scale results are much lower than that of the patch-scale results because city-scale evaluation requires longer correctly connected paths in the final graph. 

OrientationRefine presents good pixel-level performance, since it directly optimizes the results on pixels. However, the results of OrientationRefine severely suffer from topology errors, such as incorrect disconnections and ghost connections. Moreover, these topology errors could not be effectively corrected by post-processing. Thus, this method has relatively worse APLS and TLTS scores. Although enhanced-iCurb could better handle the graph topology and presents satisfactory results in most urban areas, its performance greatly drops when the scenario is complicated and irregular (e.g., in suburbs) due to the error accumulation and drifting problems. Besides, it takes a quite long time to train due to the iterative operations that are hard to accelerate. The original Sat2Graph cannot obtain meaningful results because of the isomorphic encoding issue mentioned in the last section of the Sat2Graph paper \cite{he2020sat2graph}, which makes the choice of keypoints not unique so that the graph vertices cannot be accurately extracted. Thus, in the experiment, we use graph vertices extracted by our method to implement Sat2Graph. However, different from keypoints of the road network, the road-boundary keypoints in this paper could be very far or closed to each other, which makes the encoding scheme of Sat2Graph not suitable for our task. As a result, Sat2Graph cannot effectively capture the connection information (i.e., edge) between vertices, and has inferior outcomes. Compared to the aforementioned baselines, the superiority of our csBoundary is well demonstrated. csBoundary presents good topology correctness as well as pixel-level performance without affecting the efficiency thanks to the AfANet. 

\subsection{Ablation studies}

In the ablation studies, we evaluate the necessity of local feature and global feature of the vertex embedding. The local feature captures the local visual information of graph vertices, which is critical to describe a vertex; the global feature is shared by all vertices and it represents the spatial as well as the visual information of the whole image. Both features are critical for vertex embedding, and removing either of them will harm the comprehensive description of a vertex, thus making the final evaluation results degraded. Based on the results shown in Tab. \ref{tab_ablation}, the importance and necessity of local feature and global feature of the vertex embedding are confirmed. 


\section{Conclusions and Future Work}
In this paper, we proposed csBoundary, a novel method to automatically annotate city-scale road-boundary HD maps from high-resolution aerial images. To achieve the goal, the graph of the road-boundary needs to be correctly detected. We first predicted the keypoint map and extracted graph vertices by proposed algorithms. Then inspired by the self-attention mechanism of transformer, we designed AfANet to obtain edges of the graph by predicting the adjacency matrix of graph vertices. 
CsBoundary was evaluated on a public benchmark dataset released by our previous work. Comparative experiments were conducted to verify the superiority of csBoundary over past works. We also justified the rationality of the design of csBoundary by several ablation studies. The effectiveness and efficiency of csBoundary were demonstrated by the experimental results. In the future, we plan to adapt AfANet to other line-shaped object detection tasks to illustrate the generalization ability of our proposed method.

\bibliographystyle{IEEEtran}
\bibliography{mybib}

\begin{thebibliography}{10}
\providecommand{\url}[1]{#1}
\csname url@samestyle\endcsname
\providecommand{\newblock}{\relax}
\providecommand{\bibinfo}[2]{#2}
\providecommand{\BIBentrySTDinterwordspacing}{\spaceskip=0pt\relax}
\providecommand{\BIBentryALTinterwordstretchfactor}{4}
\providecommand{\BIBentryALTinterwordspacing}{\spaceskip=\fontdimen2\font plus
\BIBentryALTinterwordstretchfactor\fontdimen3\font minus
  \fontdimen4\font\relax}
\providecommand{\BIBforeignlanguage}[2]{{%
\expandafter\ifx\csname l@#1\endcsname\relax
\typeout{** WARNING: IEEEtran.bst: No hyphenation pattern has been}%
\typeout{** loaded for the language `#1'. Using the pattern for}%
\typeout{** the default language instead.}%
\else
\language=\csname l@#1\endcsname
\fi
#2}}
\providecommand{\BIBdecl}{\relax}
\BIBdecl

\bibitem{lu2020real}
X.~Lu, Y.~Ai, and B.~Tian, ``Real-time mine road boundary detection and
  tracking for autonomous truck,'' \emph{Sensors}, vol.~20, no.~4, p. 1121,
  2020.

\bibitem{sun20193d}
P.~Sun, X.~Zhao, Z.~Xu, R.~Wang, and H.~Min, ``A 3d lidar data-based dedicated
  road boundary detection algorithm for autonomous vehicles,'' \emph{IEEE
  Access}, vol.~7, pp. 29\,623--29\,638, 2019.

\bibitem{xu2021topo}
Z.~Xu, Y.~Sun, and M.~Liu, ``Topo-boundary: A benchmark dataset on topological
  road-boundary detection using aerial images for autonomous driving,''
  \emph{IEEE Robotics and Automation Letters}, vol.~6, no.~4, pp. 7248--7255,
  2021.

\bibitem{homayounfar2018hierarchical}
N.~Homayounfar, W.-C. Ma, S.~Kowshika~Lakshmikanth, and R.~Urtasun,
  ``Hierarchical recurrent attention networks for structured online maps,'' in
  \emph{Proceedings of the IEEE Conference on Computer Vision and Pattern
  Recognition}, 2018, pp. 3417--3426.

\bibitem{homayounfar2019dagmapper}
N.~Homayounfar, W.-C. Ma, J.~Liang, X.~Wu, J.~Fan, and R.~Urtasun, ``Dagmapper:
  Learning to map by discovering lane topology,'' in \emph{Proceedings of the
  IEEE International Conference on Computer Vision}, 2019, pp. 2911--2920.

\bibitem{can2021structured}
Y.~B. Can, A.~Liniger, D.~P. Paudel, and L.~Van~Gool, ``Structured
  bird's-eye-view traffic scene understanding from onboard images,'' in
  \emph{Proceedings of the IEEE/CVF International Conference on Computer
  Vision}, 2021, pp. 15\,661--15\,670.

\bibitem{bastani2018roadtracer}
F.~Bastani, S.~He, S.~Abbar, M.~Alizadeh, H.~Balakrishnan, S.~Chawla,
  S.~Madden, and D.~DeWitt, ``Roadtracer: Automatic extraction of road networks
  from aerial images,'' in \emph{Proceedings of the IEEE Conference on Computer
  Vision and Pattern Recognition}, 2018, pp. 4720--4728.

\bibitem{tan2020vecroad}
Y.-Q. Tan, S.-H. Gao, X.-Y. Li, M.-M. Cheng, and B.~Ren, ``Vecroad: Point-based
  iterative graph exploration for road graphs extraction,'' in
  \emph{Proceedings of the IEEE/CVF Conference on Computer Vision and Pattern
  Recognition}, 2020, pp. 8910--8918.

\bibitem{he2020sat2graph}
S.~He, F.~Bastani, S.~Jagwani, M.~Alizadeh, H.~Balakrishnan, S.~Chawla, M.~M.
  Elshrif, S.~Madden, and M.~A. Sadeghi, ``Sat2graph: road graph extraction
  through graph-tensor encoding,'' in \emph{Computer Vision--ECCV 2020: 16th
  European Conference, Glasgow, UK, August 23--28, 2020, Proceedings, Part XXIV
  16}.\hskip 1em plus 0.5em minus 0.4em\relax Springer, 2020, pp. 51--67.

\bibitem{etten2020city}
A.~V. Etten, ``City-scale road extraction from satellite imagery v2: Road
  speeds and travel times,'' in \emph{Proceedings of the IEEE/CVF Winter
  Conference on Applications of Computer Vision}, 2020, pp. 1786--1795.

\bibitem{zhxu2021icurb}
Z.~Xu, Y.~Sun, and M.~Liu, ``icurb: Imitation learning-based detection of road
  curbs using aerial images for autonomous driving,'' \emph{IEEE Robotics and
  Automation Letters}, vol.~6, no.~2, pp. 1097--1104, 2021.

\bibitem{xu2021cp}
Z.~Xu, Y.~Sun, L.~Wang, and M.~Liu, ``Cp-loss: Connectivity-preserving loss for
  road curb detection in autonomous driving with aerial images,'' in \emph{2021
  IEEE/RSJ International Conference on Intelligent Robots and Systems
  (IROS)}.\hskip 1em plus 0.5em minus 0.4em\relax IEEE, 2021, pp. 1117--1123.

\bibitem{liang2019convolutional}
J.~Liang, N.~Homayounfar, W.-C. Ma, S.~Wang, and R.~Urtasun, ``Convolutional
  recurrent network for road boundary extraction,'' in \emph{Proceedings of the
  IEEE Conference on Computer Vision and Pattern Recognition}, 2019, pp.
  9512--9521.

\bibitem{mattyus2017deeproadmapper}
G.~M{\'a}ttyus, W.~Luo, and R.~Urtasun, ``Deeproadmapper: Extracting road
  topology from aerial images,'' in \emph{Proceedings of the IEEE International
  Conference on Computer Vision}, 2017, pp. 3438--3446.

\bibitem{batra2019improved}
A.~Batra, S.~Singh, G.~Pang, S.~Basu, C.~Jawahar, and M.~Paluri, ``Improved
  road connectivity by joint learning of orientation and segmentation,'' in
  \emph{Proceedings of the IEEE Conference on Computer Vision and Pattern
  Recognition}, 2019, pp. 10\,385--10\,393.

\bibitem{mi2021hdmapgen}
L.~Mi, H.~Zhao, C.~Nash, X.~Jin, J.~Gao, C.~Sun, C.~Schmid, N.~Shavit, Y.~Chai,
  and D.~Anguelov, ``Hdmapgen: A hierarchical graph generative model of high
  definition maps,'' in \emph{Proceedings of the IEEE/CVF Conference on
  Computer Vision and Pattern Recognition}, 2021, pp. 4227--4236.

\bibitem{vaswani2017attention}
A.~Vaswani, N.~Shazeer, N.~Parmar, J.~Uszkoreit, L.~Jones, A.~N. Gomez,
  {\L}.~Kaiser, and I.~Polosukhin, ``Attention is all you need,'' in
  \emph{Advances in neural information processing systems}, 2017, pp.
  5998--6008.

\bibitem{mnih2010learning}
V.~Mnih and G.~E. Hinton, ``Learning to detect roads in high-resolution aerial
  images,'' in \emph{European Conference on Computer Vision}.\hskip 1em plus
  0.5em minus 0.4em\relax Springer, 2010, pp. 210--223.

\bibitem{xue2019learning}
N.~Xue, S.~Bai, F.~Wang, G.-S. Xia, T.~Wu, and L.~Zhang, ``Learning attraction
  field representation for robust line segment detection,'' in
  \emph{Proceedings of the IEEE/CVF Conference on Computer Vision and Pattern
  Recognition}, 2019, pp. 1595--1603.

\bibitem{girard2021polygonal}
N.~Girard, D.~Smirnov, J.~Solomon, and Y.~Tarabalka, ``Polygonal building
  extraction by frame field learning,'' in \emph{Proceedings of the IEEE/CVF
  Conference on Computer Vision and Pattern Recognition}, 2021, pp. 5891--5900.

\bibitem{yun2019graph}
S.~Yun, M.~Jeong, R.~Kim, J.~Kang, and H.~J. Kim, ``Graph transformer
  networks,'' \emph{Advances in Neural Information Processing Systems},
  vol.~32, pp. 11\,983--11\,993, 2019.

\bibitem{csBoundary}
``csboundary project webpage,'' \url{https://sites.google.com/view/csboundary}.

\bibitem{lin2017feature}
T.-Y. Lin, P.~Doll{\'a}r, R.~Girshick, K.~He, B.~Hariharan, and S.~Belongie,
  ``Feature pyramid networks for object detection,'' in \emph{Proceedings of
  the IEEE conference on computer vision and pattern recognition}, 2017, pp.
  2117--2125.

\bibitem{sun2019deep}
K.~Sun, B.~Xiao, D.~Liu, and J.~Wang, ``Deep high-resolution representation
  learning for human pose estimation,'' in \emph{Proceedings of the IEEE/CVF
  Conference on Computer Vision and Pattern Recognition}, 2019, pp. 5693--5703.

\bibitem{van2018spacenet}
A.~Van~Etten, D.~Lindenbaum, and T.~M. Bacastow, ``Spacenet: A remote sensing
  dataset and challenge series,'' \emph{arXiv preprint arXiv:1807.01232}, 2018.

\bibitem{wegner2013higher}
J.~D. Wegner, J.~A. Montoya-Zegarra, and K.~Schindler, ``A higher-order crf
  model for road network extraction,'' in \emph{Proceedings of the IEEE
  Conference on Computer Vision and Pattern Recognition}, 2013, pp. 1698--1705.

\end{thebibliography}

\end{document}